\documentclass[11pt]{article}

\usepackage[final]{acl}

\usepackage{times}
\usepackage{latexsym}

\usepackage[T1]{fontenc}

\usepackage[utf8]{inputenc}

\usepackage{microtype}

\usepackage{inconsolata}

\usepackage{graphicx}
\usepackage{url}

\usepackage[dvipsnames]{xcolor}
\usepackage{pifont}
\newcommand{\tick}{\textcolor{Green}{\ding{51}}}
\newcommand{\cross}{\textcolor{Red}{\ding{55}}}
\newcommand{\cTwo}{Lavender}
\newcommand{\cThree}{SkyBlue}
\newcommand{\Goal}{\textcolor{YellowOrange}{[Goal]}}
\newcommand{\Sol}[1]{\textcolor{RoyalBlue}{[Sol#1]}}

\usepackage{makecell}
\usepackage{hhline}

\usepackage{tabularx}
\newcolumntype{Y}{>{\centering\arraybackslash}X}
\newcolumntype{Z}{>{\raggedleft\arraybackslash}X}
\usepackage{multirow}
\usepackage{fontawesome}

\title{Sinhala Physical Common Sense Reasoning Dataset for \texttt{Global PIQA}}

\author{Nisansa de Silva \\
 Dept. of Computer Science \& Engineering \\
 University of Moratuwa, 10400, Sri Lanka\\
  \texttt{NisansaDdS@cse.mrt.ac.lk} \\\And
  Surangika Ranathunga \\
 School of Math. \& Comp. Sciences\\
 Massey University, New Zealand\\
  \texttt{S.Ranathunga@massey.ac.nz} \\}

\newcommand{\entryCount}{110}

\begin{document}
\maketitle
\begin{abstract}
This paper presents the first-ever Sinhala physical common sense reasoning dataset created as part of \texttt{Global PIQA}. It contains \entryCount{} human-created and verified data samples, where each sample consists of a prompt, the corresponding correct answer, and a wrong answer. Most of the questions refer to the Sri Lankan context, where Sinhala is an official language.
\end{abstract}

\begin{center}
\raisebox{-2.2pt}{\includegraphics[scale=0.09]{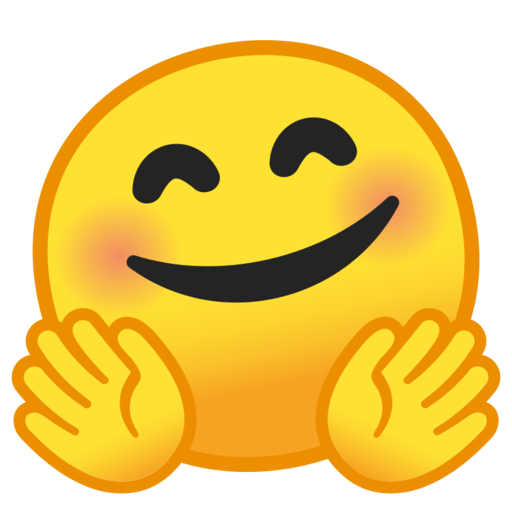}} \href{https://huggingface.co/datasets/mrlbenchmarks/global-piqa-nonparallel/viewer/sin_sinh}{~\texttt{Global PIQA (\texttt{sin\_sinh})}} 
\end{center}

\section{Sri Lankan Culture and Sinhala}
Sri Lanka is a multi-ethnic and multilingual country, with Sinhala and Tamil being the official languages. Sinhala is an Indo-Aryan language spoken by more than 17 million people in Sri Lanka~\cite{de2025survey}. It possesses a unique alphabet and script. According to~\citet{ranathunga-de-silva-2022-languages}’s language categorisation, which uses the category definition established by~\citet{joshi-etal-2020-state}, Sinhala is a low-resource language. The scarcity of digital language resources and the lack of collaborative research have been the main reasons for the slow progress of Sinhala language computing research~\cite{ranathunga-de-silva-2022-languages, de2025survey}. Given that Sri Lanka is geographically isolated from its nearest neighbour - India, Sinhala does not get included in Indian data creation initiatives or Language Models such as Indic Gemma\footnote{\url{https://dataloop.ai/library/model/telugu-llm-labs_indic-gemma-7b-finetuned-sft-navarasa-20/}}, despite being the Indo-Aryan language family.

Sri Lanka has an unbroken written history of its civilisation for more than 2500 years~\cite{geiger1908mahavamsa}. Being an island nation that has been on the busiest naval trade routes for millennia~\cite{sudharmawathie2017foreign}, it boasts a unique culture, which is both different to but also linked to other South Asian countries~\cite{roberts2021exploring}. It also had influence from its colonial rulers - Portuguese, Dutch and the British. Given that Sri Lanka is a multi-ethnic and multi-religious country, the presence of different sub-cultures is evident (e.g. Sinhala-Buddhist, Tamil-Hindu, etc).

\section{Corpus Creation}
The corpus was prepared by two Sri Lankan nationals who have lived in Sri Lanka for more than 30 years. They are the authors of this paper. Each author had their school education in Sinhala. They both hold PhDs and are NLP researchers.  Each data creator created at least 50 samples, and the other verified the quality of the samples. 

\begin{figure*}[!htb]	
	\centering 
\includegraphics[width=0.98\textwidth] 
{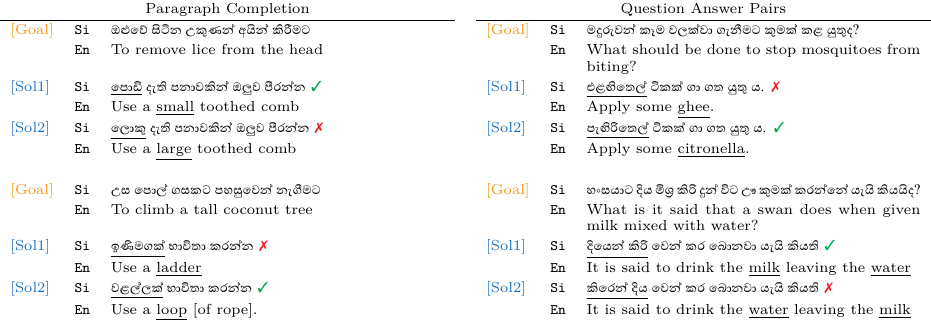}
\caption{A few examples from the Sinhala physical common sense reasoning dataset. \texttt{\Goal} is the question (prompt) while \texttt{\Sol{1}} and \texttt{\Sol{2}} are the two possible solutions, of which a human or a trained model must choose the most appropriate solution, of which exactly one is correct. The correct answer is denoted by a \tick and the incorrect answer is denoted by a \cross. Short phrases that differ between \texttt{\Sol{1}} and \texttt{\Sol{2}} are shown in \underline{underline}. In addition to the original entry in Sinhala (\texttt{Si}), in this table, we also provide an English (\texttt{En}) translation even though our dataset itself does not carry English translations.
}
\label{fig:examples}
\end{figure*}

To adhere to the conventions of the \texttt{Global PIQA} dataset~\cite{chang2025global}, this dataset follows the format of the original \texttt{PIQA} dataset~\cite{bisk2020piqa}. All the questions were manually created (i.e., they were not translated from \texttt{PIQA} or any other dataset). Some samples from the corpus are shown in Figure~\ref{fig:examples} along with the description of the notation that is used in this paper. The wrong answer for a given prompt was prepared by either 1. changing one letter of a word, 2. changing 1-3 words of a sentence, or 3. swapping words/phrases in a sentence.

\section{Corpus Statistics}

Out of the data samples, 67 are in the form of sentence/paragraph completion (Shown in the left column of Figure~\ref{fig:examples}). The remainder is in the form of Question-Answer pairs (Shown in the right column of Figure~\ref{fig:examples}).

\begin{table}[!htbp]\centering
\begin{center}
\resizebox{0.48\textwidth}{!}{

\begin{tabularx}{0.75\textwidth}{|l|Y|Y|Y|}

 \hline \multirow{2}{*}{ \textbf{Domain}} 
 & \multicolumn{3}{c|}{\textbf{Number of Samples}} \\
 \hhline{~---}
& \makecell{\textbf{Paragraph}\\\textbf{Completion}} & \makecell{\textbf{Question}\\\textbf{Answer Pairs}} & \textbf{Total}\\
 
 \hline
 Buddhism & 4& 6 & 10\\ 
 Literature & 2 & 2 & 4\\
 Mythology & 2 & 2 & 4\\
 Sports and games & 4 & 1 & 5\\
 Food & 11 & 4 & 15\\
 Farming/Agri/Fishery & 4 & 4 & 8\\
 Proverbs & 3 & 2 & 5\\
 History & 9 & 10 & 19\\
 Other & 28 & 12 & 40\\
 \hline
 \textbf{Total} & 67 & 43 & 110\\
 \hline
\end{tabularx}

}
\end{center}
\caption{Distribution of samples across different domains and types}\label{tab:stats_domains}
\end{table}

Table~\ref{tab:stats_domains} shows how the samples are distributed across different domains and the aforementioned two forms. Here, the `other' category mostly refers to common sense reasoning in day-to-day life (e.g. the first example in the left column of Figure~\ref{fig:examples}). 
Some data samples refer to general common sense applicable to any culture (e.g. how to open a tightened bottle lid?), some are very specific to the Sri Lankan culture (e.g.~ rituals related to Sri Lankan Sinhala Buddhist wedding ceremonies), while some refer to the Sri Lankan version of concepts found in other cultures (e.g. the children's game  \textit{`batto paneema'} is the Sri Lankan version of \textit{`hopscotch'}).

\begin{figure}[!hbt]	
	\centering 
\includegraphics[width=0.48\textwidth] 
{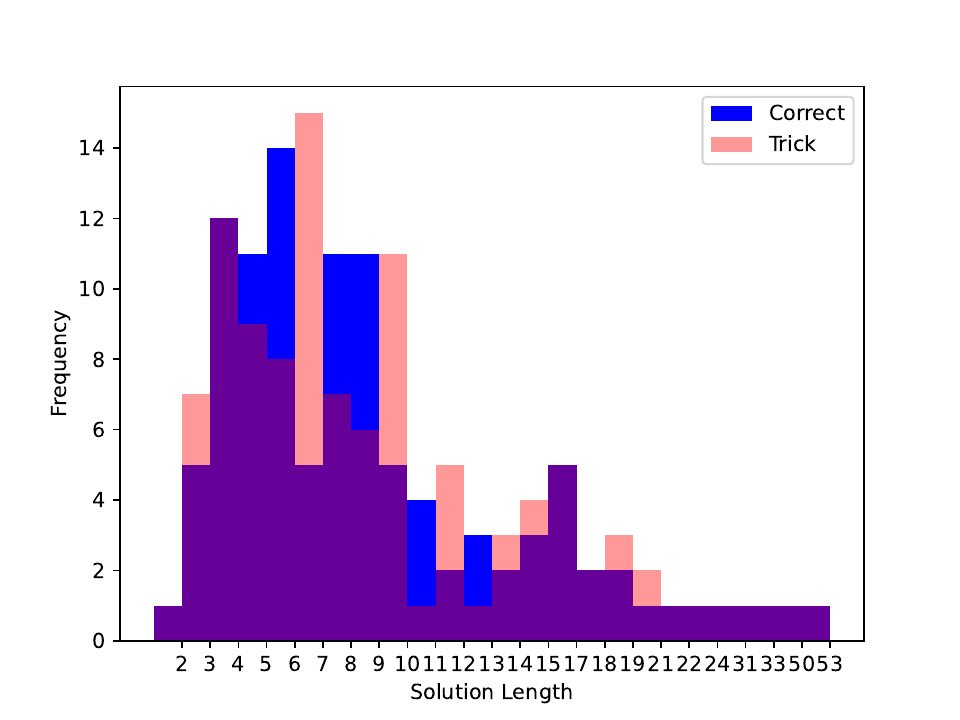}
\caption{Sentence length distributions for both correct solutions and incorrect solutions}
\label{fig:SenLen}
\end{figure}

Figure~\ref{fig:SenLen} shows a plot of the sequence lengths of the correct and incorrect solutions, as tokenised by the \texttt{SinLLaMA} tokeniser~\cite{aravinda2025sinllama}. While there are minor differences, the two distributions are reasonably similar for a dataset of \entryCount{} samples. 

We then compared the distribution of words in the data set created by us against the general domain Sinhala word distribution calculated by~\citet{wickramasinghe2023sinhala}. This yielded a Pearson correlation of 0.109404. This \textit{weak}~\cite{shi2009correlation} correlation with the general domain is expected, as this data set is specialised for the physical reasoning domain.

\section{Experiments}

\begin{figure*}[!hbt]	
	\centering 
\includegraphics[width=0.95\textwidth] 
{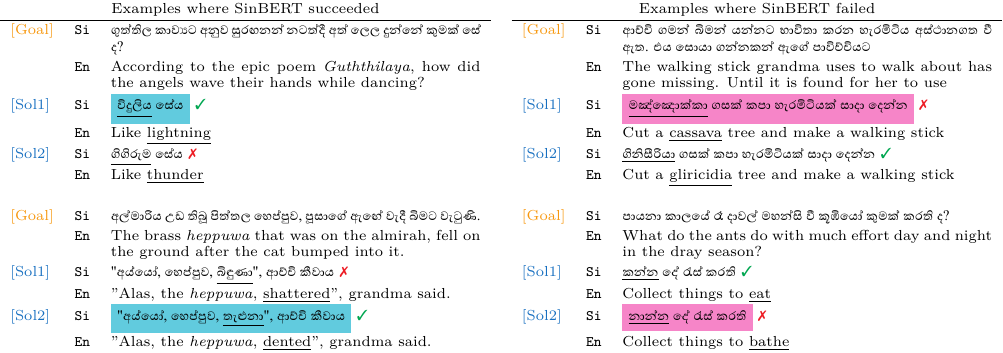}
\caption{Qualitative analysis of SinBERT’s predictions with. \textbf{Left:} Two examples that SinBERT gets right. \textbf{Right:} two examples that SinBERT gets incorrect. \colorbox{\cThree}{Blue} shading shows SinBERT succeeding while the \colorbox{\cTwo}{Red} shading shows SinBERT failing.}
\label{fig:SinBERT}
\end{figure*}

When introducing the \texttt{PIQA} dataset, \citet{bisk2020piqa} conducted an experiment on the task of \textit{Multiple Choice Question Answering} using RoBERTa~\cite{liu2019roberta} and some other language models. We also conducted an experiment of the same task using the dataset created by us. We used \texttt{SinBERT}~\cite{dhananjaya-etal-2022-bertifying} as our model, given that it is the closest Sinhala language model to the original experiment. However, one crucial difference between our experiment and the original experiment is the fact that, in the original experiment with the \texttt{PIQA} dataset, \citet{bisk2020piqa} fine-tuned the RoBERTa model and then reported the testing and validation accuracy, but given the small size (\entryCount{} entries) of our dataset, we opted for a zero-shot experiment.

\begin{table}[!htbp]\centering
\begin{center}
\resizebox{0.48\textwidth}{!}{
\begin{tabularx}{0.7\textwidth}{|l|Z|Z|Z|Z|}

 \hline \multirow{2}{*}{ \textbf{Domain}} 
 & \multicolumn{2}{c|}{\makecell[c]{\textbf{Paragraph}\\\textbf{Completion}}} & \multicolumn{2}{c|}{\makecell[c]{\textbf{Question}\\\textbf{Answer Pairs}}} \\
 \hhline{~----}
& \texttt{SinBERT} & \texttt{GPT-5} & \texttt{SinBERT} & \texttt{GPT-5}\\

 \hline
 Buddhism & 25.00 & 75.00 & 83.33 & 50.00 \\ 
 Literature & 50.00 & 100.00 & 0.00 & 50.00 \\
 Mythology & 50.00 & 100.00 & 50.00 & 0.00 \\
 Sports and games & 75.00 & 0.00 & 100.00 & 0.00 \\
 Food & 36.36 & 81.82 & 25.00 & 25.00 \\
 Farming/Agri/Fishery & 50.00 & 50.00 & 25.00 & 75.00 \\
 Proverbs & 66.67 & 100.00 & 0.00 & 50.00 \\
 History & 55.56 & 66.67 & 40.00 & 70.00 \\
 Other & 57.14 & 71.43 & 50.00 & 66.67 \\
 \hline
\end{tabularx}
}
\end{center}
\caption{Accuracy of \texttt{SinBERT} and \texttt{GPT-5} across different domains and types}\label{tab:stats_domains_SinBERT}
\end{table}

\begin{figure}[!hbt]	
	\centering 
\includegraphics[width=0.49\textwidth] 
{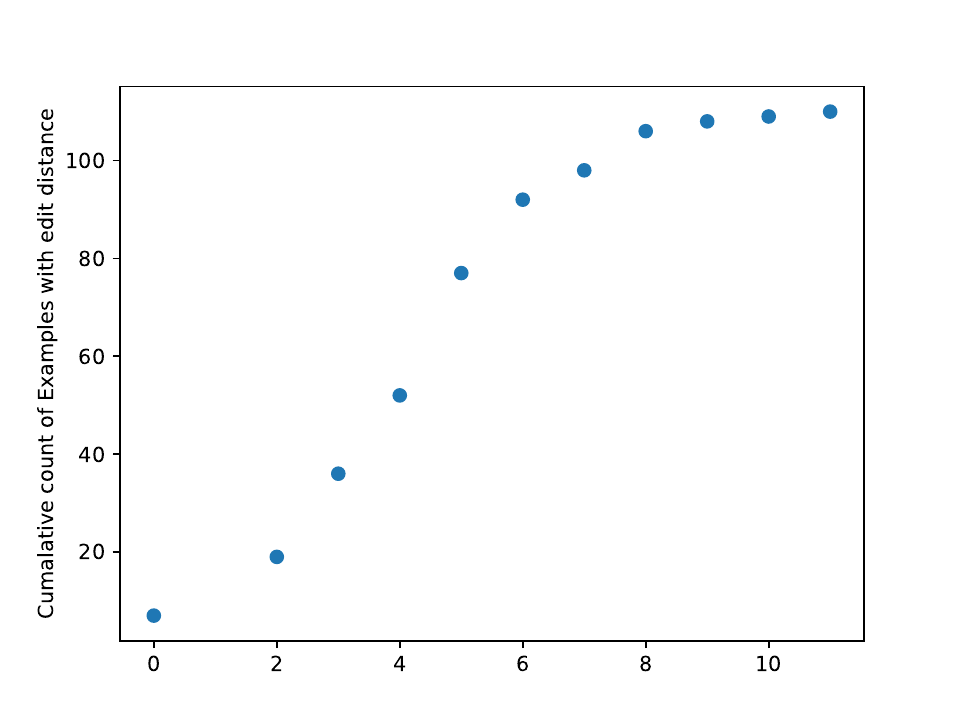}
\includegraphics[width=0.49\textwidth] 
{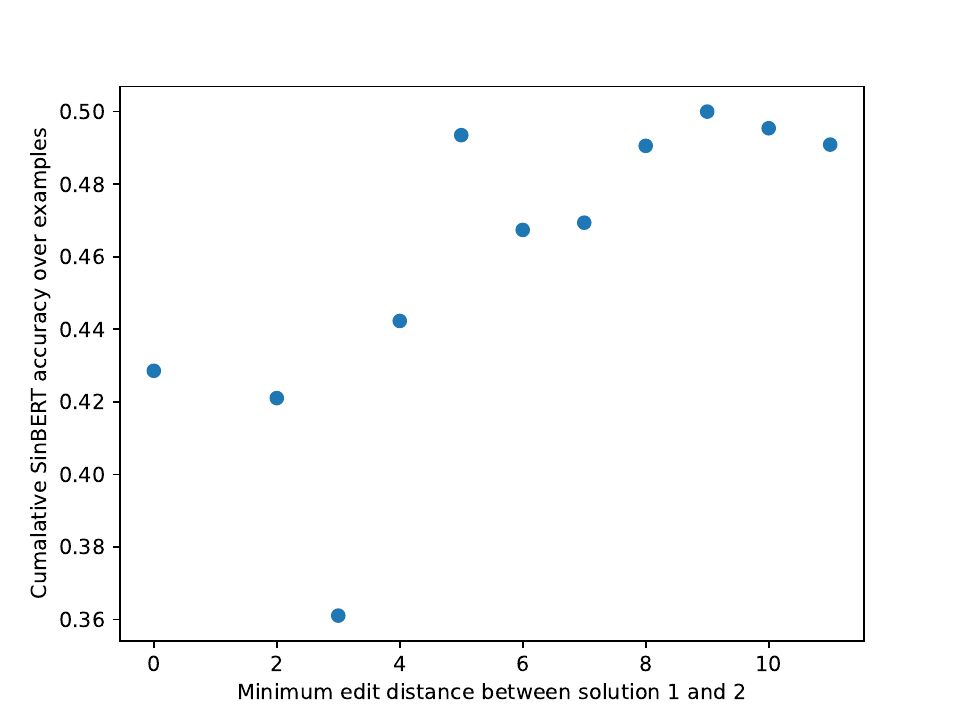}
\caption{Breaking down the data set by edit distance between solution choices. \textbf{Top:} Cumulative histogram of examples in the data set, in terms of minimum edit
distance d between the two solution choices. The majority of the dataset consists of small tweaks between the two solution pairs; nevertheless, this is enough to confuse NLP models. \textbf{Bottom:} Zero-Shot SinBERT accuracy over examples with a minimum edit distance. Dataset difficulty increases somewhat as the two solution pairs are allowed to drift further apart.}
\label{fig:SenEditDist}
\end{figure}

The accuracy achieved by \texttt{SinBERT} across different domains and sample types is shown in Table~\ref{tab:stats_domains_SinBERT}. In most domains, \texttt{SinBERT} seems to do better when the sample is stated as an instance of \textit{paragraph completion} rather than a direct \textit{question answer pair}. 
Further, we show some of the qualitative analysis of this experiment in Figure~\ref{fig:SinBERT}.

\begin{figure}[!hbt]	
	\centering 
\fbox{\includegraphics[width=0.48\textwidth] 
{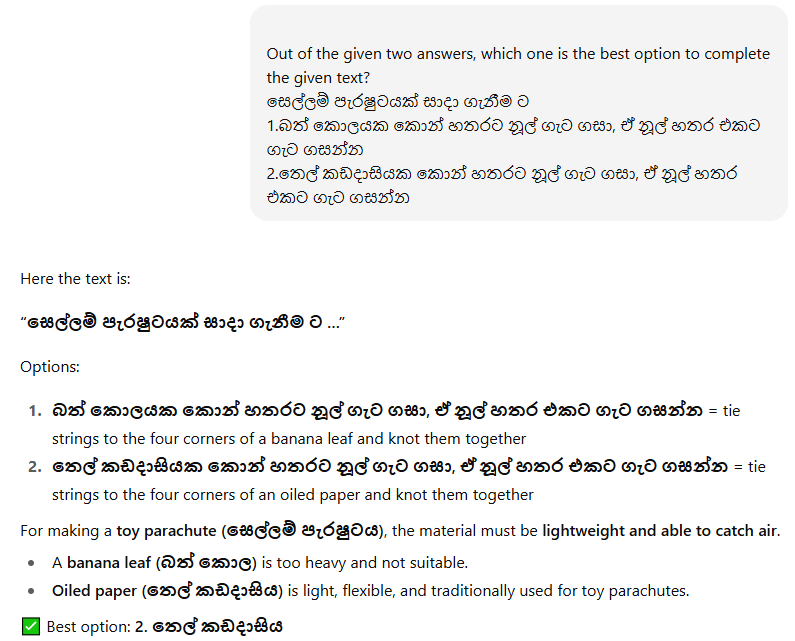}}
\fbox{\includegraphics[width=0.48\textwidth] 
{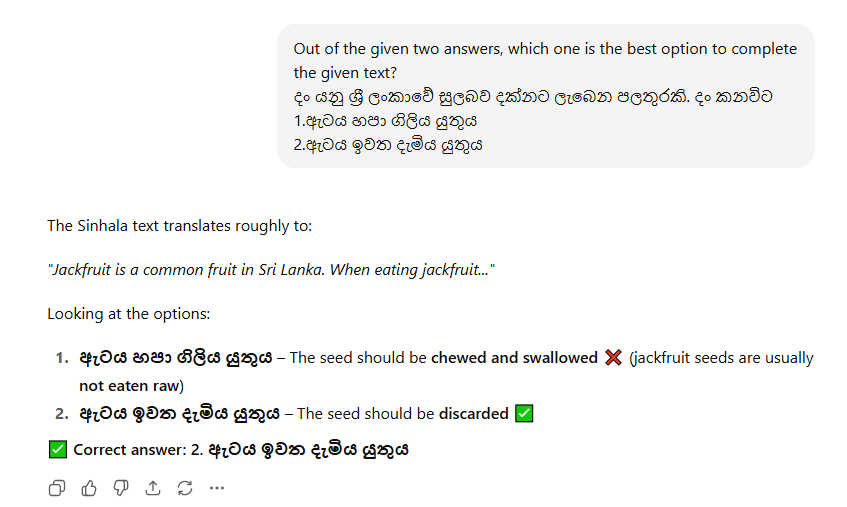}}
\caption{Two sample outputs from GPT-5 mini.}
\label{fig:gpt}
\end{figure}
What we can observe from this is that while \texttt{SinBERT} may have some success on the examples where general common sense might reasonably arrive at the correct answer, the deeper the question and answers are embedded into the Sri Lankan culture, the more it starts to struggle to provide answers. In fact, the overall accuracy landed at a dismal $49.09$, which is even lower than the expected value of a uniform random guess.

In Figure~\ref{fig:SenEditDist} we show the distribution of the examples against the edit distance between the two solution choices. We compute edit distance over tokenised strings with a cost of 1 for edits, insertions, and deletions. We also show how the zero-shot results of \texttt{SinBERT}~\cite{dhananjaya-etal-2022-bertifying} change over the edit distance.

We also tested GPT-5 mini free version (Accessed on 17/09/2025) with the \entryCount{} data samples. It managed to answer 71/110 ($64.5\%$) samples correctly.
The accuracy achieved by \texttt{GPT-5} across different domains and sample types is shown in Table~\ref{tab:stats_domains_SinBERT}. In most domains, \texttt{GPT-5} seems to do better than \texttt{SinBERT}. However, specifically in \textit{question answer pair} type under \textit{Buddhism}, \textit{Sports and games} and \textit{Mythology} domains \texttt{SinBERT} succeeds over \texttt{GPT-5}.

We noted that GPT-5 mini translates the Sinhala text to English before reasoning on it. Sometimes, translation errors result in the model providing wrong answers. One example is shown in the top part of Figure~\ref{fig:gpt}. The Sinhala text \textit{`bath kolaya'} refers to a thin polythene that is used to wrap rice to make a packet of rice.  GPT-5 mini incorrectly translates this term to \textit{banana leaf}. This is a very interesting case, because before polythene was introduced to Sri Lanka, banana leaves were primarily used to wrap rice. However, in Sinhala, a banana leaf is refereed to as \textit{`kesel kolaya}', while \textit{`bath kolaya'} refers to the polythene. Due to this wrong translation, GPT-5 mini produces the wrong answer to the question. Interestingly, the opposite happens in the second (bottom) example of Figure~\ref{fig:gpt}. It refers to \textit{`dan}' \raisebox{-0.5ex}{ 
\includegraphics[height=1.4\fontcharht\font`\A]{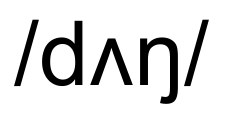} } (\textit{Syzygium caryophyllatum (L.) Alston}), a wild fruit endemic to Sri Lanka. It is the size of a blue berry. While eating it, the seed has to be thrown away. However, GPT-5 mini translates \textit{`dan}' to the jackfruit seed, which is not eaten raw. The final answer of GPT-5 mini results in the correct answer, because of this wrong translation, despite GPT-5 not knowing about \textit{dan}.

Results of the further experiments conducted by~\citet{chang2025global} in assembling the \texttt{Global PIQA} dataset are available at \faGithub~\href{https://github.com/mrlbenchmarks/global-piqa/tree/main/eval_results}{\texttt{mrlbenchmarks}}.

\section{Conclusion}
This paper presented the first-ever physical common sense reasoning dataset written in Sinhala. We evaluated two models - SinBERT and GPT-5 with this dataset. While GPT-5 outperformed, its overall accuracy is just $64.5\%$. This underscores the challenges faced by modern-day Large Language Models, when working with data from different cultures, written in languages other than English.

\section*{Limitations}

Given that both data creators are Sinhala Buddhists, the dataset has a bias towards Sinhala Buddhist culture. There can be minor spelling mistakes in the dataset (while there has been some research on implementing Sinhala spell correctors, none is at a production-ready level~\cite{sonnadara2021sinhala, gunathilake2025lmspell}). We tested GPT-5 mini only once via its GUI interface. Due to its non-deterministic nature, it may produce different results if tested multiple times with the same question.

\bibliography{anthology,custom}

\end{document}